\documentclass[conference]{IEEEtran}

\usepackage{graphicx}
\usepackage{times}
\usepackage{latexsym}
\usepackage{epsfig}
\usepackage{epstopdf}
\usepackage{tikz}
\usepackage{amssymb, amsmath}
\usepackage{amsfonts}
\usepackage{listings}
\usepackage{url}
\usepackage{mathtools}

\begin{document}
\title{About Learning in\\Recurrent Bistable Gradient Networks}

\author{\IEEEauthorblockN{J\"orn Fischer}
\IEEEauthorblockA{
Mannheim University\\of Applied Sciences\\
Paul-Wittsack-Str. 10\\
68163 Mannheim\\
Germany\\
Tel.: (+49) 621 292-6767\\
Fax: (+49) 621 292-6-6767-1\\
Email: j.fischer@hs-mannheim.de\\
}

\and
\IEEEauthorblockN{Steffen Lackner}
\IEEEauthorblockA{Mannheim University\\of Applied Sciences\\
Paul-Wittsack-Str. 10\\
68163 Mannheim\\
Email: steffen.lackner@posteo.de}
}

\maketitle

\begin{abstract}
Recurrent Bistable Gradient Networks~\cite{chinarov00,chinarov01,chinarov03} are attractor based neural networks characterized by bistable dynamics of each single neuron. Coupled together using linear interaction determined by the interconnection weights, these networks do not suffer from spurious states or very limited capacity anymore. Vladimir Chinarov and Michael Menzinger, who invented these networks, trained them using Hebb's learning rule. We show, that this way of computing the weights leads to unwanted behaviour and limitations of the networks capabilities.
Furthermore we evince, that using the first order of Hintons Contrastive Divergence $CD^{1}$ algorithm~\cite{hinton95} leads to a quite promising recurrent neural network. These findings are tested by learning images of the MNIST database for handwritten numbers.
\end{abstract}

\IEEEpeerreviewmaketitle{}

\section{Introduction}
Hopfield networks invented in 1984 by John Hopfield~\cite{hopfield82, hopfield84} are somehow predecessors of Deep Belief Networks which are widely used as state of the art neural networks.
They are recurrent neural networks inspired by the physical behaviour of spin glasses. Hopfield networks are perceptron based and have a symmetric weight matrix and no self-connecting neurons. This guarantees that all dynamics, that can take place in this type of network, is a fixed point attraction.
To overcome negative effects as spurious states and limited capacity of Hopfield Networks, Bolzmann Machines~\cite{hinton86} were introduced which were then restricted to have no interconnections between neurons in a layer and which were stacked and trained layer by layer e.g.\ with the Wake Sleep Algorithm introduced by Geoffrey Hinton~\cite{hinton95}.
In the BioSystems journal~\cite{chinarov00} in the year 2000 and on the IWANN conference in 2001~\cite{chinarov01} Vladimir Chinarov and Michael Menzinger presented a class of recurrent Hopfield-like networks called Bistable Gradient Networks, which eliminated the disadvantages of spurious states and of the very limited capacity. They demonstrated this by training these networks successfully with the Hebbian learning rule, storing many more patterns than a standard Hopfield network could memorize ($k = 0.14 N$, with $N$ neurons).

Because of their successful implementation of $5$ interconnected neurons, their paper~\cite{chinarov01} presents Hebb's rule as the perfect, efficient way to train the Bistable Gradient Networks.
In our investigation we realize, that this is not always true. There are pattern combinations which may not be stored with Hebb's learning rule.
In the following section we start with a short description of the basic principles of Bistable Gradient Networks.
To understand why Hebb's learning rule is not the best choice to train them, a simple thought experiment is described afterwards.
We show, that using Hintons Contrastive Divergence $CD^1$ leads to far better results. Furthermore we demonstrate the capabilities storing handwritten numbers from the MNIST-database into the network and point out that noisy images are nearly perfectly denoised. Finally we end up with a conclusion.

\section{Bistable Gradient Networks}
In this section a short introduction to the basic concepts of Bistable Gradient Networks is given.
In the domain of dynamical systems a neuron is written down as a differential equation. To derive this equation we start with the energy function of such a neuron, which may be defined as follows:

\begin{equation}
\begin{aligned}
V(x_i) &= V_{bistable} + V_{couple}\\
       &= (- \frac{x_{i}^{2}}{2}+\frac{x_{i}^{4}}{4}) + (-\frac{1}{2} \sum_{i} \sum_{j} w_{ij} x_{i} x_{j}),
\end{aligned}
\end{equation}
where $V_{bistable}$ leads to a bistable behaviour of the neuron and $V_{couple}$ describes the linear coupling between the neurons.

The variable $x$ defines the neurons state or output, while the derivative of the energy function with respect to $x$ gives us the direction, in which the neurons state $x$ changes in time:

\begin{equation}
\frac{dx_i}{dt} = -\frac{\partial V(x_i)}{\partial x_i} =  (x_i-x_{i}^{3}) + (\sum_{j}w_{ij}x_{j})
\label{difequ}
\end{equation}

This energy function or potential is shown in figure~\ref{Potential}; the derivative is plotted below in figure~\ref{Derivative}. The minima of the energy function correspond to the fixed points marked in the figure of its derivative.
In the differential equation~(\ref{difequ}) we can see that there is a linear part---the sum of the weighted outputs---which may shift the function $x-x^{3}$ up or down as shown in figure~\ref{Derivative} as a dashed line. In dependence of this linear part it easily happens that only the left or only the right fixed point exists anymore. This leads to a predetermined behaviour. The neural output converges to $1$ (or slightly above) or to $-1$ or (slightly below).
Let us name the state $x_i=1$ active and $x_i=-1$ inactive.
If a number of neurons have positive interconnection weights and a large part of these neurons is active, then their derivative $x_j-x_j^{3}$ will be shifted up and the inactive neurons converge to the active state. On the other hand a neuron which is active, but connected with negative weights to and from the other active neurons, will shift its derivative down and it converges to the inactive state.

\begin{figure}[htb]
\centering
\includegraphics[width=3in]{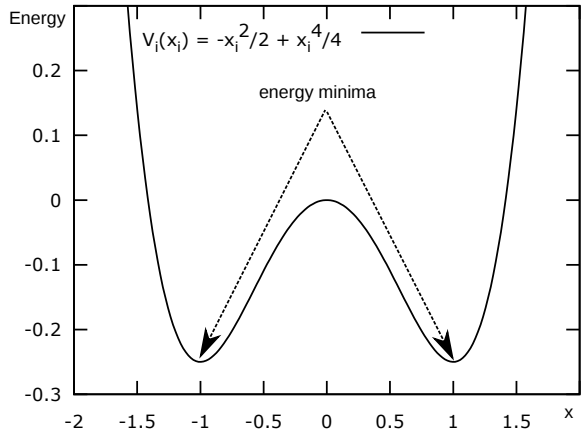}
\caption{The energy function $V(x_i)$ has two minima, where the activity of the neuron converges to. This energy function explains the bistability of the neurons.}
\label{Potential}{}
\end{figure}

\begin{figure}[htb]
\centering
\includegraphics[width=3in]{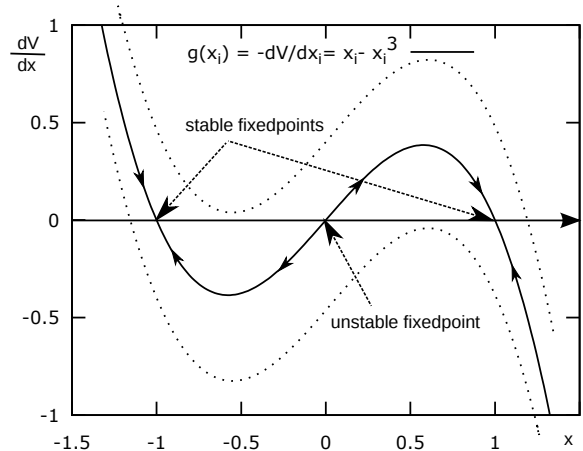}
    \caption{The stable fixed points are related to the energy minima in figure~\ref{Potential}. The derivative is zero, if $x=1$ (active) or $x=-1$ (inactive). The dashed curves show how the derivative is shifted up or down in dependence of the weights and the output of other neurons according to~(\ref{difequ}). If it is shifted up, only the positive fixed point exists, while if it is shifted down only the negative fixed point remains.}
\label{Derivative}
\end{figure}

\section{Thought Experiment}
To understand why special pattern combinations may not be stored we first write down the Hebbian learning rule:

\begin{equation}
w_{ij} = \frac{1}{p} \beta \sum_k{x_i(k) x_j(k) }
\end{equation}

Especially if we store only one pattern it is easily seen, that the active and inactive neurons get strong positive interconnections and the connections between active and inactive neurons will be strongly negative.
This implicates, that the inverse image is always stored as strong as the image itself into the network, a phenomenon which is also described by Hopfield~\cite{hopfield82,hopfield84}.

If we now try to store many patterns into a network, which strongly overlap e.g.\ a big number of active neurons for all patterns with a small number of neurons which make the difference between the patterns, we find a problem emerging:
a big number of neurons which is always active will always inhibit a small area of mostly inactive neurons, even if a few of them are active for a stored pattern. The network would always activate the big area, while the rest would be certainly always deactivated.

In the following section we describe how to change the learning rule to observe the wanted behaviour.

\section{Contrastive Divergence $CD^1$}

Though the type of neurons Geoffrey Hinton uses are completely different (he uses binary output with stochastic activation) the learning rule is of great interest for us.
The learning rule for $CD^n$ may be written down as follows:

\begin{equation}
w_{ij}^{\tau+1} = w_{ij}^{\tau} + \eta {(<x_{i}x_{j}>_0 - <x_{i}x_{j}>_n)},
\end{equation}

where $\tau$ denotes the time step. $CD^{1}$ is computed as:

\begin{equation}
w_{ij}^{\tau+1} = w_{ij}^{\tau} + \eta {(<x_{i}x_{j}>_0 - <x_{i}x_{j}>_1)}
\label{cd1}
\end{equation}

We start with randomly initialized weights $w_{ij}\leq 0.01$ $\forall i,j \in [1..N]$. We initialize the networks output with the pattern to be learned. $<x_{i}x_{j}>_0$ is calculated from this initialisation. After computing the activation of the network for one time step we receive $<x_{i}x_{j}>_1$. To adapt the weights only~(\ref{cd1}) has to be applied for all patterns for several times.

If a pattern is represented by a fixed point the difference in~(\ref{cd1}) $(<x_{i}x_{j}>_0 - <x_{i}x_{j}>_1)$  will be zero and the weights stay unchanged. If a neuron changes its state after activation, the difference $(<x_{i}x_{j}>_0 - <x_{i}x_{j}>_1)$ may be positive or negative.
In the case of a negative difference the weights are weakened, while if it is positive the weights are strengthened. This is done until the difference for all patterns is zero, so that each pattern to be learned results in a fixed point.

After each weight change we keep the neurons connections symmetric $w_{ji} = w_{ij}$ and eliminate self connections $W_{ii}=0$. These two conditions guarantee our network to contain only fixed point attractors. This is because any state change will decrease the appropriate energy function. In further experiments we neglected these constraints and see that the behavior of the network does not change remarkably.

In the next section the algorithm is tested on patterns of the MNIST-database for handwritten numbers.

\enlargethispage{-2.0in}

\section{The MNIST-database for handwritten numbers}

\begin{figure}[htb]
\centering
\includegraphics[width=3.5in]{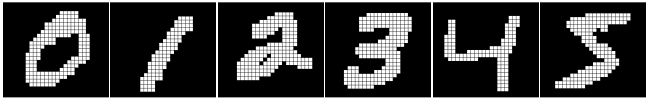}
\caption{The upper images show the letters which are trained. The learning rate is $\eta=0.01$. Each pattern is trained alternatively for $50$ iterations.}
\label{learn}
\end{figure}

To proof that learning with the $CD^1$ algorithm is successful, we trained a network of $28\times28$ neurons with patterns from the MNIST-database using~(\ref{cd1}). An excerpt of these learning patterns is shown in figure~\ref{learn}. The great overlap of the neurons activation from one handwritten number to another makes it impossible to train these patterns using the Hebbian learning rule.

\begin{figure}[htb]
\centering
\includegraphics[width=3.5in]{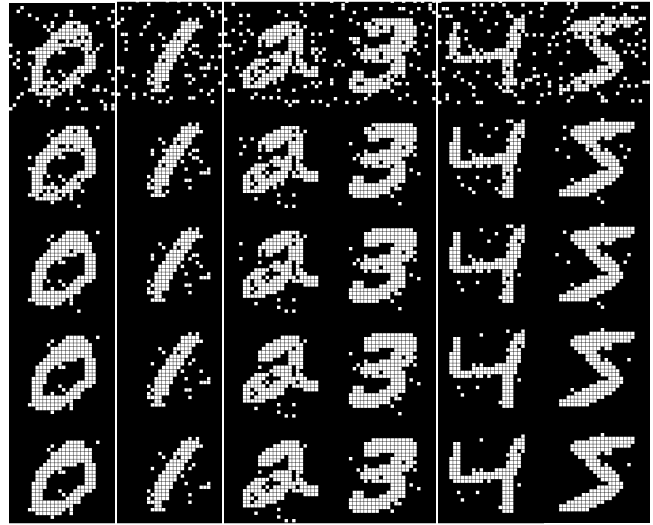}
\caption{After training, we start with a noisy $28\times28$ pixel image, where $100$ pixel at a random position are inverted. The images below are computed using the Euler-method with a time step size of $0.1$ in $25$ iterations. Only on each $5$-th iteration an image is presented.}

\label{test}
\end{figure}

The handwritten numbers are trained for $50$ iterations with a learning rate $\eta = 0.01$. Figure~\ref{test} shows the reconstruction of the original images out of images with more than $10\%$ of noise added. The network is activated using the Euler-method with a step size of $0.1$. Each $5$-th time step an image is computed. The converged images have a mean error rate about $1.4 \%$.

\section{Conclusion}

The recurrent Bistable Gradient Network using Hebb's learning rule for computing the interconnection weights of the network leads to difficulties especially in strongly overlapping patterns. To overcome these problems we applied Hintons first order Contrastive Divergence $CD^1$ algorithm to train the weights. The results were successfully tested with patterns from the MNIST-database for handwritten letters. Testing an image reconstruction with noisy images of more than $10\%$ of noise leads to a near perfect reconstruction with a mean error rate of about $1.4 \%$. In our future research we will try to improve learning by taking higher orders of the $CD^n$ algorithm into account.

\newpage

\section*{Acknowledgment}
The authors would like to thank Tobias Becht for helpful comments.


\begin{thebibliography}{1}

\bibitem{chinarov00}
V. Chinarov, M. Menzinger, Computational dynamics of gradient bistable networks, BioSystems 55, p 137-142, 2000

\bibitem{chinarov01}
V. Chinarov, M. Menzinger, Bistable Gradient Neural Networks: Their Computational Properties, IWANN2001 Conference in Granada, Spain, proceedings pp 333-338, Springer, 2001

\bibitem{chinarov03}
V. Chinarov, M. Menzinger, Reconstruction of noisy patterns by bistable gradient neural like networks, BioSystems 68, p 147-153, 2003

\bibitem{hinton95}
G. Hinton, Hinton, P. Dayan,, B. Frey., and R. Neal. The wake-sleep algorithm for
self-organizing neural networks. Science, 268, 1158–1161, 1995

\bibitem{hinton86}
G. E. Hinton, T. J. Sejnowski, D. E. Rumelhart, J. L. McClelland, PDP Research Group, Learning and Relearning in Boltzmann Machines, Parallel Distributed Processing: Explorations in the Microstructure of Cognition. Volume 1: Foundations. Cambridge: MIT Press: 282–317, 1986

\bibitem{hopfield82}
J. J. Hopfield, Neural networks and physical systems with emergent collective computational properties. Proc. Nat. Acad. Sci. (USA) 79, 2554-2558., 1982
\bibitem{hopfield84}
J. J. Hopfield, Neurons with graded response have collective computational properties like those of two-state neurons. Proc. Nat. Acad. Sci. (USA) 81, 3088-3092.,1984


\end{thebibliography}
\end{document}